\RequirePackage{fix-cm}
\documentclass[smallextended]{svjour3}       % onecolumn (second format)
\smartqed  % flush right qed marks, e.g. at end of proof

\usepackage{graphicx}
\usepackage{amsmath}
\usepackage{amssymb}
\usepackage{algorithmic}
\usepackage{algorithm}
\usepackage{subfigure}
\usepackage{array}
\usepackage{tabularx}
\usepackage{multirow}
\usepackage[pagewise]{lineno}

\newcommand{\bff}{{\textbf{f}}}

\newcommand{\bfz}{{\textbf{z}}}
\newcommand{\bfh}{{\textbf{h}}}

\newcommand{\bfx}{{\textbf{x}}}

\newcommand{\bfu}{{\textbf{u}}}

\newcommand{\bfg}{{\textbf{g}}}

\newcommand{\bfalpha}{{\boldsymbol{\alpha}}}
\newcommand{\bfbeta}{{\boldsymbol{\beta}}}

\begin{document}

\title{Stochastic Learning of Multi-Instance Dictionary  for Earth Mover's Distance based Histogram Comparison}

\author{Jihong Fan \and Ru-Ze Liang}

\institute{Jihong Fan \at
Qiqihar Medical University, Qiqihar 161006, Heilongjiang, People's Republic of China\\
\email{fanjihong43@yahoo.com}\\
\texttt{Jihong Fan is the corresponding author.}
\and
Ru-Ze Liang\at
King Abdullah University of Science and Technology, Saudi Arabia\\
\email{ruzeliang@outlook.com}
}

\date{Received: date / Accepted: date}

\maketitle

\begin{abstract}
Dictionary plays an important role in multi-instance data representation. It maps bags of instances to histograms. Earth mover's distance (EMD) is the most effective histogram distance metric for the application of multi-instance retrieval. However, up to now, there is no existing multi-instance dictionary learning methods designed for EMD based histogram comparison. To fill this gap, we develop the first EMD-optimal dictionary learning method using stochastic optimization method. In the stochastic learning framework, we have one triplet of bags, including one basic bag, one positive bag, and one negative bag. These bags are mapped to histograms using a multi-instance dictionary. We argue that the EMD between the basic histogram and the positive histogram should be smaller than that between the basic histogram and the negative histogram. Base on this condition, we design a hinge loss. By minimizing this hinge loss and some regularization terms of the dictionary, we update the dictionary instances. The experiments over multi-instance retrieval applications shows its effectiveness when compared to other dictionary learning methods over the problems of medical image retrieval and natural language relation classification.
\end{abstract}

\section{Introduction}

In the problem of multi-instance learning, the data is given as bags of instances, i.e., each data sample has multiple instances. An example of multi-instance data is the medical image data. In medical imaging problem, an medical image is usually divided to small regions, and each region is a instance of this image. Thus when the image is processed, we call the image a bag of instances \cite{Yan20161332,Li20151727,Vanwinckelen2016313,Chen2016587,Yan20161332}. An another example is the audio single date. In the problem of audio single processing problem, a sequence of audio single is usually split into a set of short frames, and each frame is regard as a instance, while the sequence itself is treated as a bag of instances \cite{ma2014design,chen2009sub,chen2011design,chen2012low,ju20121}. We investigate the problem of retrieving multi-instance bags from a database. The inessential components are the representation of the bag, and the calculation of the similarity/ dissimilarity between two bags of instances. Currently, the most popular method to represent a bag of instances to use a multi-instance dictionary. The instances of a bag is mapped to the dictionary, and the similarity between the instances of the bag and the dictionary is calculated and normalized to a histogram. The most important part of this representation method is to learn a multi-instance dictionary \cite{Huo201276,Huo2014,Wu2015428}. To calculate the dissimilarity between two bags, we can calculate a distance between two histograms. Currently, existing works have show the most effective distance metric to compare a pair of histograms is earth mover's distance (EMD) \cite{Nabavi2016963,Zhang20161,Beecks2016233}. EMD treats the bins of a histogram as a set of supplies of earth, and the bins of another histogram as a set of demands of earth. It moves the earth from the bins of supplies to the bins of demands. The minimize amounts of moved earth weighted by some ground distances is calculated as EMD. Compared to some other histogram distance metric, EMD has the ability to cross-bin similarities, thus yields better comparison results.

Currently, there are many existing works proposed to learn the multi-instance dictionary. The simplest method is to perform a $k$-clustering to a set of training instances \cite{Ying20154258,Kim2016198,Xu201665}. Some more complex learning methods use the dictionary to represent bags, and minimize the classification errors of the bags to learn the dictionary \cite{chen2006miles,fu2011milis,Lobel20152218,wang2013max}. The disadvantages of these methods are of two folds when it is applied to the retrieval problem based on EMD.

\begin{enumerate}
\item The existing methods of dictionary learning ignores the EMD comparison of histograms. A dictionary which is optimal to the minimization of classification errors is not necessarily optimal for the retrieval problem based on EMD distance metric. Up to now, there is no theoretical proof or experimental study to show that a classification error minimization based dictionary learning method can obtain a EMD-comparison based bag-histogram comparison.

\item The existing methods use all the training bags simultaneously to optimize the dictionary. If the size of the training set is large, the optimization process can be very time-consuming. In some cases, it is even unacceptable when the big data is considered.
\end{enumerate}

To solve these problems, we propose to learn an optimal multi-instance dictionary specifically for the EMD based histogram comparison. We also propose that the optimization of this learning process should be efficient, and we choose the stochastic optimization method. This method does not use all the training bags simultaneously, but use the training bags one by one in an iterative algorithm. In each iteration, only one training bag is input to the algorithm, and thus avoids the time-consuming process of processing all the training bags.

The proposed method takes a triplets of multi-instance bags, which are one basic bag, one positive bag and one negative bag. The positive bag belongs to the same class as the basic bag, but the negative bag belongs to a different class from the basic bag. The three bags are represented by a multi-instance dictionary and normalized to three histograms. Then we use the EMD distance metric to calculate the dissimilarity between the basic histogram and the negative histogram, and the dissimilarity between the basic histogram and the positive histogram. Moreover, we argue that the first distance should be larger than the seconde distance plus a margin amount, since we hope the distance to the negative histogram is bigger than that to the positive histogram. We also define a hinge loss to learn the dictionary when this condition does not hold.  Moreover, we derive that the EMD between two histograms is a linear weighted combination of the bins of the histograms. We learn the dictionary by updating the dictionary instances to minimize the hinge loss and the squared $\ell_2$ norm of the dictionary instances.

The following parts of this paper is organized in the following forms. In section \ref{sec:back} we introduce the backgrounds of this work, including the multi-instance dictionary representation, the EMD, and the stochastic optimization framework. In section \ref{sec:method}, the proposed dictionary learning method is modeled, optimized, and an iterative algorithm is derived. In section \ref{sec:exp}, the proposed method is evaluated in the application of multi-instance bag retrieval.  In section \ref{sec:conclu}, the conclusion of this paper is given.

\section{Backgrounds}
\label{sec:back}

In this section, we give some brief introduction to the relevant background knowledge of our work. Our work is a novel multi-instance dictionary learning method for EMD metric, and it is based on the maximization of large margins and stochastic learning technology. Thus we introduce the backgrounds of the multi-instance dictionary learning, the EMD metric, the large margin learning, and the stochastic learning.

\subsection{Multi-instance data representation using dictionary}

Suppose we have a pattern recognition problem, and the input data is multi-instance data. Each data sample is given as a bag of multiple instances, $\mathcal{B}=\{\bfx_i\}_{i=1}^{n}$, where $\bfx_i \in \mathbb{R}^\phi$ is $\phi$-dimensional feature vector of the $i$-th instance of the bag, and $n$ is the number of instances of the bag. For example, in the problem of image representation, we can use the bag-of-words method to represent an image. The image is split to many small patches, and each patch is treated as an instance, while the image is a bag of instances. To represent a bag of instances, we can use a multi-instance dictionary. The dictionary is a set of $m$ instances, denoted as $\mathcal{D}=\{\bfz_j\}_{j=1}^{m}$, where $\bfz_j$ is the $\phi$-dimensional feature vector of $j$-th dictionary instance. Then we can quantize the bag of instances to the dictionary, and use the quantization histogram as the bag-level feature vector of the bag. To obtain the histogram, we calculate the normalized similarities between the bag and dictionary instances, and concatenate them,

\begin{equation}
\begin{aligned}
h(\mathcal{B}; \mathcal{D})=[s(\mathcal{B},\bfz_1), \cdots, s(\mathcal{B},\bfz_m)]^\top,
\end{aligned}
\end{equation}
where $h(\mathcal{B}; \mathcal{D})$ is the quantization histogram of $\mathcal{B}$ given $\mathcal{D}$ as the dictionary, and $s(\mathcal{B},\bfz_j)$ is the normalized similarity between $\mathcal{B}$ and the $j$-th dictionary instance $\bfz_j$. To calculate the normalized similarity $s(\mathcal{B},\bfz_j)$, we first calculate the original similarity $\overline{s}(\mathcal{B},\bfz_j)$ as the summation of the instance similarities between the instances of $\mathcal{B}$ and $\bfz_j$ measured by a Gaussian kernel function,

\begin{equation}
\begin{aligned}
\overline{s}(\mathcal{B},\bfz_j) = \sum_{i=1}^n \exp \left ( - \frac{\|\bfx_i - \bfz_j\|_2^2}{2\sigma^2} \right )
\end{aligned}
\end{equation}
where $\sigma$ is the bandwidth parameter of the kernel function. Beside the summation measure of the similarities, we may also use the maximum measure or some other measures. The normalized similarity $s(\mathcal{B},\bfz_j)$ is defined using the original similarity measures as

\begin{equation}
\begin{aligned}
s(\mathcal{B},\bfz_j) = \frac{\overline{s}(\mathcal{B},\bfz_j)}{\sum_{j'=1}^m \overline{s}(\mathcal{B},\bfz_j')}
=
\frac{1}{\pi}
\sum_{i=1}^n \exp \left ( - \frac{\|\bfx_i - \bfz_j\|_2^2}{2\sigma^2} \right ),
\end{aligned}
\end{equation}
where $\pi = \sum_{j'=1}^m \overline{s}(\mathcal{B},\bfz_j')$ is the normalization term. Using the bag-level features, we can train classifiers to predict class labels. For example, Chen et al. \cite{chen2006miles} and Fu et al. \cite{fu2011milis} proposed to represent the bags to bag-level features and train support vector machine for classification problems.

\subsection{Histogram comparison using earth mover's distance}

In many applications, we need to compare the dissimilarity between two histograms. For example, in the nearest neighbor classification problem, we need to search the nearest neighbors of a test data sample from the database first, and then assign the labels of the neighbors to the test point. The most effective distance measure to compare a pair of histograms is the EMD. Suppose we have two histograms $\bfh = [h_1, \cdots, h_m]^\top$ and $\bfg = [g_1,\cdots,g_{m'}]^\top$ of $m$ bins, where $h_j$ and $g_j$ are the $j$-th bins of $\bfh$ and $\bfg$ respectively. To calculate the EMD between $\bfh$ and $\bfg$, we consider the bins of $\bfh$ as supplies of earth, and bins of $\bfg$ as demands of earth. We move earth from supplies to the demands. The distance of moving a unit of earth from $h_j$ to $g_k$ is denoted as $d_{jk}$, and the amount of earth moved from $h_j$ to $g_k$ is denoted as $f_{jk}$. The overall EMD of moving the earth of $\bfh$ to $\bfg$ is given as the summation of the moved earth weighted by the bin-to-bin distances,

\begin{equation}
\begin{aligned}
D(\bfh,\bfg) = \sum_{j,k=1}^m f_{jk} d_{jk}.
\end{aligned}
\end{equation}
Some constraints are imposed to the amounts of moved earth,

\begin{equation}
\begin{aligned}
&\sum_{k=1}^m f_{jk} = h_j, \forall~j=1,\cdots,m,\\
&\sum_{j=1}^m f_{jk} = g_k, \forall~k=1,\cdots,m,\\
&\sum_{j,k=1}^m f_{jk} = 1, ~and,~f_{jk}\geq 0 ,\forall~j,k=1,\cdots,m.
\end{aligned}
\end{equation}
These constrains define a search space for the amounts of earth, and the final EMD is the minimum overall moved earth with regard to the amounts of earth in this space. To obtain the EMD, we need to solve the following minimization problem to solve the amounts of bin-to-bin moving of earth,

\begin{equation}
\begin{aligned}
f_{jk}|_{j,k=1}^m = \underset{f'_{jk}|_{j,k=1}^m} {\arg\min}
~&\sum_{j,k=1}^m f'_{jk} d_{jk},\\
st.~&\sum_{k=1}^m f'_{jk} = h_j, \forall~j=1,\cdots,m,\\
&\sum_{j=1}^m f'_{jk} = g_k, \forall~k=1,\cdots,m,~and,\\
&f'_{jk}\geq 0 ,\forall~j,k=1,\cdots,m.
\end{aligned}
\end{equation}
This is the constrained linear programming problem.

\subsection{Stochastic large margin learning}

For many learning problems, the training data is large, thus the training process can be very time-time consuming. One way to solve this problem is to using the stochastic learning strategy. In this strategy, instead of using all the training data samples simultaneously, we use the training data samples one by one. Recently, a novel stochastic optimization method is proposed to learn multi-kernel similarity function \cite{Xia2014536}. Each input is a triplet of data samples $\mathcal{Q} = (p,p^+,p^-)$, composed of a base data sample $p$, a positive data sample $p^+$ which is from the same class as $p$, and a negative data sample $p^-$ which is from a different class from $p$. Given a parametric similarity function $S(\cdot,\cdot)$, it is argued that the similarity between $p$ and $p^+$ should be larger than that between $p$ and $p^-$ plus a marginal amount of one,

\begin{equation}
\begin{aligned}
S(p,p^+) > S(p,p^-) + 1.
\end{aligned}
\end{equation}
The corresponding hinge loss function of this argument is

\begin{equation}
\begin{aligned}
\ell(p,p^+,p^-) = \max (0,  S(p,p^-) + 1 - S(p,p^+)).
\end{aligned}
\end{equation}
In an iterative algorithm, in current iteration with the a training triplet $\mathcal{Q}$, the stochastic optimization problem of updating $S$ is given as follows,

\begin{equation}
\begin{aligned}
\min_{S} \left \{ \frac{1}{2} \|S-S^{pre}\|^2_2 + C\ell(p,p^+,p^-) \right \},
\end{aligned}
\end{equation}
where $S^{pre}$ is the solution of $S$ of previous iteration, and $\frac{1}{2} \|S-S^{pre}\|^2_2 $ is minimized to respect the previous solution. $\ell(p,p^+,p^-)$ is minimized to utilize the current training triplet. $C$ is a tradeoff parameter between the first term and the second term.

\section{Proposed Method}
\label{sec:method}

In this section, we will introduce a novel stochastic learning method of multi-instance dictionary for EMD measure. To represent multi-instance bags, we proposed to learn a multi-instance dictionary, $\mathcal{D}=\{\bfz_j\}_{j=1}^m$. To this end, we use a stochastic learning strategy in an iterative algorithm. In the $t$-th iteration, we have a training triplet of bags, $\mathcal{Q} = (\mathcal{B},\mathcal{B}^+,\mathcal{B}^-)$, where $\mathcal{B}=\{\bfx_i\}_{i=1}^n$ is a base bag, $\bfx_i$ is its $i$-th instance, and $n$ is the number of instances of $\mathcal{B}$. $\mathcal{B}^+=\{\bfx_j^+\}_{i=1}^{n_+}$ is a positive bag, it belongs to the same class as $\mathcal{B}$, $\bfx_i^+$ is its $i$-th instance, and $n_+$ is the number of instances of the positive bag. $\mathcal{B}^-=\{\bfx_j^-\}_{i=1}^{n_-}$ is a negative bag, it belongs to a different class from $\mathcal{B}$, $\bfx_i^-$ is its $i$-th instance, and $n_-$ is the number of instances of the negative bag. Using the multi-instance dictionary, $\mathcal{D}$ we represent the bags of $\mathcal{Q}$ as histograms of quantization,

\begin{equation}
\begin{aligned}
&\bfh = [h_1, \cdots,h_m]^\top,\\
&\bfg = [g_1, \cdots,g_m]^\top, and\\
&\bfu = [u_1, \cdots,u_m]^\top,\\
\end{aligned}
\end{equation}
where $\bfh$ is the histogram of $\mathcal{B}$, $\bfg$ is the histogram of $\mathcal{B}^+$, and $\bfu$ is the histogram of $\mathcal{B}^-$. $h_j$ is the $j$-th bin of $\bfh$, and it is defined as

\begin{equation}
\begin{aligned}
&h_j = \frac{1}{\pi} \sum_{i=1}^n \exp \left ( -\frac{\|\bfx_i-\bfz_j\|_2^2}{2\sigma^2}\right ),~where\\
&\pi = \sum_{j'=1}^m \sum_{i=1}^n \exp \left ( -\frac{\|\bfx_i-\bfz_{j'}\|_2^2}{2\sigma^2}\right ).
\end{aligned}
\end{equation}
$g_i$ and $u_i$ are defined in similar ways.

To compare the distance between $\mathcal{B}$ and $\mathcal{B}^+$, we calculate the EMD between their histograms, $\bfh$ and $\bfg$. The EMD between them is defined as,

\begin{equation}
\label{equ:Z}
\begin{aligned}
&Z = \min_{f_{jk},j,k=1,\cdots,m}\sum_{j,k=1}^m f_{jk} d_{jk}, \\
&s.t.~\sum_{k=1}^m f_{jk} = h_j,\forall~j=1,\cdots,m,\\
&\sum_{j=1}^m f_{jk} = g_k, \forall~k=1,\cdots,m,~and\\
&f_{jk} \geq 0,  \forall~j,k=1,\cdots,m.
\end{aligned}
\end{equation}
where $f_{jk}$ is the amount of earth moved from $j$-th bin of $\bfh$, $h_j$, to the $k$-th bin of the $\bfg$, $g_k$. We define a moved earth amount matrix, $F = [f_{jk}] \in R^{m\times m}$, with its $(j,k)$-th element $f_{jk}$, and a vector representation of the matrix $\bff = vec(F) \in R^{m^2}$, where $vec(F)$ is an operator converting the matrix $F$ into a vector by concatenating the columns of the matrix to one long column. Similarly, we also define a distance matrix $D = [d_{jk}] \in \mathbb{R}^{m\times m}$, and its corresponding vector $vec(D)$. In this wey, we can rewrite $Z$ as follows,

\begin{equation}
\begin{aligned}
&Z=\sum_{j,k=1}^m f_{jk} d_{jk}=Tr(F^\top D) = vec(D)^\top vec(F),\\
&Z -  vec(D)^\top vec(F) = 0.
\end{aligned}
\end{equation}
We also define a matrix for each of the constraints in (\ref{equ:Z}). For the constraint $\sum_{k=1}^m f_{jk} = h_j$, we define $\Theta_j\in \{1,0\}^{m\times m}$ with only its $j$-th row containing elements of ones, while all other elements zeros, so that,

\begin{equation}
\label{equ:hconstr}
\begin{aligned}
\sum_{k=1}^m f_{jk} = Tr(\Theta_j^\top F)=vec(\Theta_j)^\top vec(F) = h_j, \forall j=1,\cdots,m.
\end{aligned}
\end{equation}
Similarly, we define $\Psi_k \in \{1,0\}^{m\times m}$ with only its $k$-th column containing elements of ones, while all other elements zeros, and

\begin{equation}
\label{equ:gcosntr}
\begin{aligned}
\sum_{j=1}^m f_{jk} = Tr(\Psi_k ^\top F) = vec(\Psi_k)^\top vec(F) = g_k, \forall~k=1,\cdots,m.
\end{aligned}
\end{equation}
Combining constraints of (\ref{equ:hconstr}) and (\ref{equ:gcosntr}) into matrix form, we have

\begin{equation}
\label{equ:constraint}
\begin{aligned}
&
\begin{bmatrix}
vec(\Theta_1)^\top \\
\vdots \\
vec(\Theta_m)^\top \\
vec(\Psi_1)^\top \\
\vdots \\
vec(\Psi_m)^\top
\end{bmatrix}
vec(F)
=
\begin{bmatrix}
h_1 \\
\vdots \\
h_m \\
g_1 \\
\vdots \\
g_m
\end{bmatrix},
\\
&
\Omega~ vec(F) = \bfalpha,
\end{aligned}
\end{equation}
where $\Omega = \begin{bmatrix}
vec(\Theta_1)^\top \\
\vdots \\
vec(\Theta_m)^\top \\
vec(\Psi_1)^\top \\
\vdots \\
vec(\Psi_m)^\top
\end{bmatrix}$, and $\bfalpha = \begin{bmatrix}
h_1 \\
\vdots \\
h_m \\
g_1 \\
\vdots \\
g_m
\end{bmatrix}$. In (\ref{equ:constraint}), there are $m^2$ variables, but only $2m$ constraints. Thus we have only $2m$ basic variables, and $m^2 - 2m$ nonbasic variables. We split the vector $vec(F)$ into vector of basic variables, $vec(F)_B$, and vector of nonbasic variables, $vec(F)_{NB}$. $\Omega$ and $vec(D)$ is also split in a similar way, and we have $\Omega_B$, $\Omega_{NB}$, $vec(D)_B$, $vec(D)_{NB}$ and

\begin{equation}
\label{equ:constraint1}
\begin{aligned}
&\left [\Omega_B~\Omega_{NB}\right ]
\begin{bmatrix}
vec(F)_B\\
vec(F)_{NB}
\end{bmatrix}
 =
\bfalpha
\Rightarrow
\left [0~\Omega_B~\Omega_{NB}\right ]
\begin{bmatrix}
Z\\
vec(F)_B\\
vec(F)_{NB}
\end{bmatrix}
 =
\begin{bmatrix}
0\\
\bfalpha
\end{bmatrix}
,\\
&
[1~-vec(D)_B^\top~-vec(D)_{NB}^\top]
\begin{bmatrix}
Z\\
vec(F)_B\\
vec(F)_{NB}
\end{bmatrix}
=
\begin{bmatrix}
0\\
\bfalpha
\end{bmatrix}.
\end{aligned}
\end{equation}
We further combine the first and second lines of (\ref{equ:constraint1}) to one single equation of matrices as follows,

\begin{equation}
\label{equ:constraint2}
\begin{aligned}
&
\begin{bmatrix}
&1&-vec(D)_B^\top&-vec(D)_{NB}^\top\\
&0&\Omega_B&\Omega_{NB}
\end{bmatrix}
\begin{bmatrix}
Z\\
vec(F)_B\\
vec(F)_{NB}
\end{bmatrix}
=
\begin{bmatrix}
0\\
\bfalpha
\end{bmatrix}\\
&\Rightarrow
\begin{bmatrix}
&1&-vec(D)_B^\top\\
&0&\Omega_B
\end{bmatrix}^{-1}
\begin{bmatrix}
&1&-vec(D)_B^\top&-vec(D)_{NB}^\top\\
&0&\Omega_B&\Omega_{NB}
\end{bmatrix}
\begin{bmatrix}
Z\\
vec(F)_B\\
vec(F)_{NB}
\end{bmatrix}\\
&
=
\begin{bmatrix}
&1&-vec(D)_B^\top\\
&0&\Omega_B
\end{bmatrix}^{-1}
\begin{bmatrix}
0\\
\bfalpha
\end{bmatrix}.
\end{aligned}
\end{equation}
We further have

\begin{equation}
\label{equ:H}
\begin{aligned}
\begin{bmatrix}
&1&-vec(D)_B^\top\\
&0&\Omega_B
\end{bmatrix}^{-1}
=
\begin{bmatrix}
&1&vec(D)_B^\top\Omega_B^{-1}\\
&0&\Omega_B^{-1}.
\end{bmatrix}
\end{aligned}
\end{equation}
Substituting (\ref{equ:H}) back to (\ref{equ:constraint2}), we have

\begin{equation}
\label{equ:constraint3}
\begin{aligned}
&
\begin{bmatrix}
&1&vec(D)_B^\top\Omega_B^{-1}\\
&0&\Omega_B^{-1}
\end{bmatrix}
\begin{bmatrix}
&1&-vec(D)_B^\top&-vec(D)_{NB}^\top\\
&0&\Omega_B&\Omega_{NB}
\end{bmatrix}
\begin{bmatrix}
Z\\
vec(F)_B\\
vec(F)_{NB}
\end{bmatrix}\\
&
=
\begin{bmatrix}
&1&vec(D)_B^\top\Omega_B^{-1}\\
&0&\Omega_B^{-1}
\end{bmatrix}
\begin{bmatrix}
0\\
\bfalpha
\end{bmatrix}\\
&\Rightarrow
\begin{bmatrix}
&1&0&-vec(D)_{NB}^\top + vec(D)_B^\top\Omega_B^{-1} \Omega_{NB}\\
&0&I&\Omega_B^{-1} \Omega_{NB}
\end{bmatrix}
\begin{bmatrix}
Z\\
vec(F)_B\\
vec(F)_{NB}
\end{bmatrix}\\
&
=
\begin{bmatrix}
vec(D)_B^\top\Omega_B^{-1}\bfalpha\\
\Omega_B^{-1}\bfalpha
\end{bmatrix}\\
&
\Rightarrow
Z- \left [ -vec(D)_{NB}^\top + vec(D)_B^\top\Omega_B^{-1} \Omega_{NB} \right ] vec(F)_{NB} = vec(D)_B^\top\Omega_B^{-1}\bfalpha
\end{aligned}
\end{equation}
Please note that the optimal solution of $F$ will obtained when $vec(F)_{NB} = \bf0$, thus we can rewrite (\ref{equ:constraint3}) as

\begin{equation}
\label{equ:Zsolution}
\begin{aligned}
Z
&= vec(D)_B^\top\Omega_B^{-1}\bfalpha\\
&= vec(D)_B^\top\Omega_B^{-1} [h_1,\cdots, h_m, g_1, \cdots, g_m]^\top \\
&=
\bfbeta [h_1,\cdots, h_m, g_1, \cdots, g_m]^\top\\
&=\sum_{j=1}^m \beta_j h_{j} + \sum_{k=1}^m \beta_{m+k} g_k
\end{aligned}
\end{equation}
where $\bfbeta = vec(D)_B^\top\Omega_B^{-1} = [\beta_1,\cdots,\beta_{2m}] \in \mathbb{R}^{2m} $.
In a similar way, we can also have the EMD between $\bfh$ and $\bfu$ as

\begin{equation}
\label{equ:Zsolution}
\begin{aligned}
\Gamma = \sum_{j=1}^m \beta_j h_{j} + \sum_{k=1}^m \beta_{m+k} u_k
\end{aligned}
\end{equation}
To learn the optimal dictionary, we proposed that the EDM between the histograms of $\mathcal{B}$ and $\mathcal{B}^-$, $\Gamma$, should be larger than the EDM between the histograms of $\mathcal{B}$ and $\mathcal{B}^+$, $Z$, plus a margin amount $\tau$

\begin{equation}
\label{equ:conditiion}
\begin{aligned}
&\Gamma > Z+\tau\\
&\Rightarrow \sum_{j=1}^m \beta_j h_{j} + \sum_{k=1}^m \beta_{m+k} u_k> \sum_{j=1}^m \beta_j h_{j} + \sum_{k=1}^m \beta_{m+k} g_k
+\tau\\
&\Rightarrow \sum_{k=1}^m \beta_{m+k} u_k> \sum_{k=1}^m \beta_{m+k} g_k
+\tau\\
&\Rightarrow 0> \sum_{k=1}^m \beta_{m+k} \left (g_k - u_k\right )
+\tau
\end{aligned}
\end{equation}
Its corresponding loss function is

\begin{equation}
\label{equ:conditiion}
\begin{aligned}
\ell(\bfg,\bfu)
&= \max\left (0,\sum_{k=1}^m \beta_{m+k} \left (g_k - u_k\right ) + \tau \right )\\
&= \max\left (0, \sum_{k=1}^m \beta_{m+k} \left (
\frac{1}{\pi^+} \sum_{i=1}^{n^+} \exp \left ( -\frac{\|\bfx_i^+-\bfz_k\|_2^2}{2\sigma^2}\right )
\right. \right.\\
&
\left.\left.
 -
\frac{1}{\pi^-} \sum_{i=1}^{n^-} \exp \left ( -\frac{\|\bfx_i^- -\bfz_k\|_2^2}{2\sigma^2}\right )
 \right ) + \tau \right )\\
&= \xi \left (\sum_{k=1}^m \beta_{m+k} \left (
\frac{1}{\pi^+} \sum_{i=1}^{n^+} \exp \left ( -\frac{\|\bfx_i^+-\bfz_k\|_2^2}{2\sigma^2}\right )
\right. \right.\\
&
\left.\left.
 -
\frac{1}{\pi^-} \sum_{i=1}^{n^-} \exp \left ( -\frac{\|\bfx_i^- -\bfz_k\|_2^2}{2\sigma^2}\right )
 \right ) + \tau \right ),
\end{aligned}
\end{equation}
where $\xi$ is defined as

\begin{equation}
\label{equ:xi}
\begin{aligned}
\xi = 1,~ if~\sum_{k=1}^m \beta_{m+k} \left (g_k - u_k\right ) + \tau>0,~ and~0~otherwise.
\end{aligned}
\end{equation}
To learn the optimal dictionary, we propose to minimize this loss function regard to the dictionary.

\textbf{Remark}: Please note that in our work, we use the triplets of histograms as training data. Each triplet contains a basic histogram, a positive histogram, and a negative histogram. But in the final objective, the basic histogram has vanished in (\ref{equ:conditiion}). The loss function is only the function of the bins of the positive and negative.

Moreover, we also propose to make solution of each dictionary instance as simple as possible, and to minimize the squared $\ell_2$ norm of $\bfz_j$. Thus the minimization problem of the updating of the dictionary instances is

\begin{equation}
\label{equ:objective}
\begin{aligned}
{\min}_{\bfz_j|_{j=1}^m}&
\left \{O(\bfz_1,\cdots,\bfz_m) =
\frac{1}{2} \sum_{j=1}^m \|\bfz_j\|_2^2
\right.
\\
&
+ C
\xi \left ( \sum_{k=1}^m \beta_{m+k} \left (
\frac{1}{\pi^+} \sum_{i=1}^{n^+} \exp \left ( -\frac{\|\bfx_i^+-\bfz_k\|_2^2}{2\sigma^2}\right )
\right. \right.
\\
& \left. \left.\left.
 -
\frac{1}{\pi^-} \sum_{i=1}^{n^-} \exp \left ( -\frac{\|\bfx_i^- -\bfz_k\|_2^2}{2\sigma^2}\right )
 \right ) + \tau \right ) \right \}.
\end{aligned}
\end{equation}
To minimize this problem, we first fix $\bfz_j|_{j=1}^m$ to update $\xi$ and $\pi^+$ and $\pi^-$, and then use the gradient descent algorithm to update $\bfz_j$ one by one. The sub-gradient function of $O$ with regard to $\bfz_k$ is

\begin{equation}
\label{equ:gradient}
\begin{aligned}
&
\nabla_{\bfz_k} O(\bfz_k) =
\bfz_k
\\
&
+ C \xi  \beta_{m+k} \left (
\frac{1}{\pi^+} \sum_{i=1}^{n^+} \exp \left ( -\frac{\|\bfx_i^+-\bfz_k\|_2^2}{2\sigma^2}\right )
\frac{(\bfx_i^+-\bfz_k)}{\sigma^2}
\right.
\\
& \left.
 -
\frac{1}{\pi^-} \sum_{i=1}^{n^-} \exp \left ( -\frac{\|\bfx_i^- -\bfz_k\|_2^2}{2\sigma^2}\right )
\frac{(\bfx_i^- -\bfz_k)}{\sigma^2}
 \right ).
\end{aligned}
\end{equation}
The updating rule of $\bfz_k$ is

\begin{equation}
\label{equ:update}
\begin{aligned}
\bfz_k^{new} \leftarrow \bfz_k^{old} - \eta \nabla_{\bfz_k} O(\bfz_k^{old}),
\end{aligned}
\end{equation}
where $\eta$ is the descent step, and its value is decided by cross-validation. Using this stochastic updating rule, we can design an iterative algorithm to learn the dictionary for EMD based histogram comparison as in Algorithm \ref{alg:overall}.

\begin{algorithm}[!htb]
\caption{Iterative algorithm for stochastic learning of the multi-instance for EMD based histogram comparison (SD-EMD).}
\label{alg:overall}
\begin{algorithmic}

\STATE Initialize $\bfz_j|_{j=1}^m$;

\REPEAT

\STATE In put one new training triplet, $(\mathcal{B},\mathcal{B}^+,\mathcal{B}^-)$.

\STATE Update $\pi^+$ and $\pi^-$,

\begin{equation}
\begin{aligned}
&\pi^+ = \sum_{j'=1}^m \sum_{i=1}^{n^+} \exp \left ( -\frac{\|\bfx_i^+-\bfz_{j'}\|_2^2}{2\sigma^2}\right ),~
and~
\pi^- = \sum_{j'=1}^m \sum_{i=1}^{n^-} \exp \left ( -\frac{\|\bfx_i^--\bfz_{j'}\|_2^2}{2\sigma^2}\right ).
\end{aligned}
\end{equation}

\STATE Update $g_k$ and $u_k$ for $k=1,\cdots,m$,

\begin{equation}
\begin{aligned}
&g_k = \frac{1}{\pi^+} \sum_{i=1}^{n^+} \exp \left ( -\frac{\|\bfx_i^+-\bfz_{k}\|_2^2}{2\sigma^2}\right ),~
and~
u_k = \frac{1}{\pi^-}\sum_{i=1}^{n^-} \exp \left ( -\frac{\|\bfx_i^--\bfz_{k}\|_2^2}{2\sigma^2}\right ).
\end{aligned}
\end{equation}

\STATE Update $\xi$,

\begin{equation}
\begin{aligned}
\xi = 1,~ if~\sum_{k=1}^m \beta_{m+k} \left (g_k - u_k\right ) + \tau>0,~ and~0~otherwise.
\end{aligned}
\end{equation}

\IF{$\xi\neq 0$}

\STATE Calculate the sub-gradient function for $k=1,\cdots,m$,

\begin{equation}
\label{equ:gradient}
\begin{aligned}
&
\nabla_{\bfz_k} O(\bfz_k) =
\bfz_k
\\
&
+ C \xi  \beta_{m+k} \left (
\frac{1}{\pi^+} \sum_{i=1}^{n^+} \exp \left ( -\frac{\|\bfx_i^+-\bfz_k\|_2^2}{2\sigma^2}\right )
\frac{(\bfx_i^+-\bfz_k)}{\sigma^2}
\right.
\\
& \left.
 -
\frac{1}{\pi^-} \sum_{i=1}^{n^-} \exp \left ( -\frac{\|\bfx_i^- -\bfz_k\|_2^2}{2\sigma^2}\right )
\frac{(\bfx_i^- -\bfz_k)}{\sigma^2}
 \right ).
\end{aligned}
\end{equation}

\ENDIF

\STATE Update $\bfz_k$ for $k=1,\cdots,m$,

\begin{equation}
\begin{aligned}
\bfz_k \leftarrow \bfz_k - \eta \nabla_{\bfz_k} O(\bfz_k).
\end{aligned}
\end{equation}

\UNTIL{All training triplets are processed.}

\STATE \textbf{Output}: $\mathcal{D}= \{\bfz_k\}|_{k=1}^m$.
\end{algorithmic}
\end{algorithm}

\section{Experiments}
\label{sec:exp}

In the experiments, we evaluate the proposed method for the problem of multi-instance data retrieval.

\subsection{Data sets}

We use three data sets in our experiment. They are the Digital Database for Screening Mammography (DDSM) \cite{Rose2006376}, and the SemEval-2010 Task 8 data set \cite{hendrickx2009semeval}.

In the DDSM data set \cite{Rose2006376}, there are 2620 cases of three classes, which is normal, cancer, and benign. Each case has two images of screening mammography, which is the images of the left breast and the right breast. To represent each case, we divide the two images to small regions and extract SIFT and pixel features from each region. Thus each case contains a number of regions, and each region is treated as an instance.

The SemEval-2010 Task 8 data set is a data set for relation classification of words \cite{hendrickx2009semeval}. In this data set, each data sample is a sentence, and in this sentence, two works are tagged as label words. The problem is to predict the relation between these two words. This dat set contains 10,717 examples of 9 different classes of relation types, which are Content-Container, Cause-Effect,
Component-Whole,  Entity-Origin, Entity-Destination, Instrument-Agency, Message-Topic, Member-Collection, and Product-Producer. Moreover, one more relation type is also considered, which is the Other type. Thus there are totally ten types. Each example contains multiple works, and we treat each work as an instance, thus it is a multi-instance data classification problem. We represent the words using its word embedding features, and its position embedding with regard to the two target words.

\subsection{Experimental process}

Given a data set, we use the ten-fold cross-validation experimental protocol to conduct the experiment \cite{Tong20162965,Shao2016260}. The entire data set is split into ten folds with the same size, and we use each of them as a query set in turn. The remaining nine folds are combined and used as a database set. Given a query data sample, we want to retrieve the data samples of a database of the same class. The retrieval is based on EMD-based comparison. We first represent each data sample as a histogram, and then calculate the EMD between the query histogram and each histogram of the data samples of a database.

\subsection{Performance measures}

To evaluate the retrieval performance, we use the recall-precision curve \cite{Goadrich2006231,Zhang20121348}, and the receiver  operating  characteristic curve \cite{Clarkson2016930,Wang20161907}. Given a query, the retrieval system returns a number of database samples. The number of data samples which are in the returned list and also belong to the same class as the query is defined as true positive. The number of the other data samples in the returned list is defined as false positive. The number of data samples which belong to the same class as the query but not in the returned list is defined as the false negative. The recall and precision is defined accordingly,

\begin{equation}
\begin{aligned}
&recall = \frac{true~positive}{true~positive+fales~negative},~and\\
&precision = \frac{true~positive}{true~positive+false+positive}.
\end{aligned}
\end{equation}
If we change the number of the returned database samples, we will have different pairs of recall and precision values. By plotting these pairs in one single figure, we can obtain the recall-precision curve. A good retrieval system can give a recall-precision curve close to the top-right corner of the figure.

In the returned list of a query, the number of samples which are from different classes from the query is defined as the true negative. The true positive rate and the false positive rate are defined as follows,

\begin{equation}
\begin{aligned}
&true~positive~rate = \frac{true~positive}{true~positive+fales~negative},~and\\
&false~positive~rate = \frac{false~positive}{false~positive+true+negative}.
\end{aligned}
\end{equation}
Similar to the recall-precision curve, we can also obtain different pairs of true~positive~rate and false~positive~rate by changing the size of returned list. Also, by plotting these pairs in one single figure, we can have the receiver  operating  characteristic curve, and a receiver  operating  characteristic curve which is close to the top-left corner of the figure is preferred.

\subsection{Experimental results}

In the experiments, we compare the proposed method with some other multi-instance dictionary learning methods. To make the comparison fair, we use these dictionary learning methods to learn the dictionaries over the database set, and then represent the data samples as histograms. The histograms are compared by using the EMD and ranked. The compared dictionary learning methods are listed as follows,

\begin{itemize}
\item max-margin multiple-instance dictionary learning (MMD) \cite{wang2013max},
\item domain transfer multi-instance dictionary learning (DTD) \cite{wang2016domain},
\item multiple-instance learning via embedded instance selection (MILES) \cite{chen2006miles}, and
\item multiple instance learning with instance selection (MILIS) \cite{fu2011milis}.
\end{itemize}

\subsubsection{Results over the DDSM data set}

\begin{figure}
\centering
\includegraphics[width=0.7\textwidth]{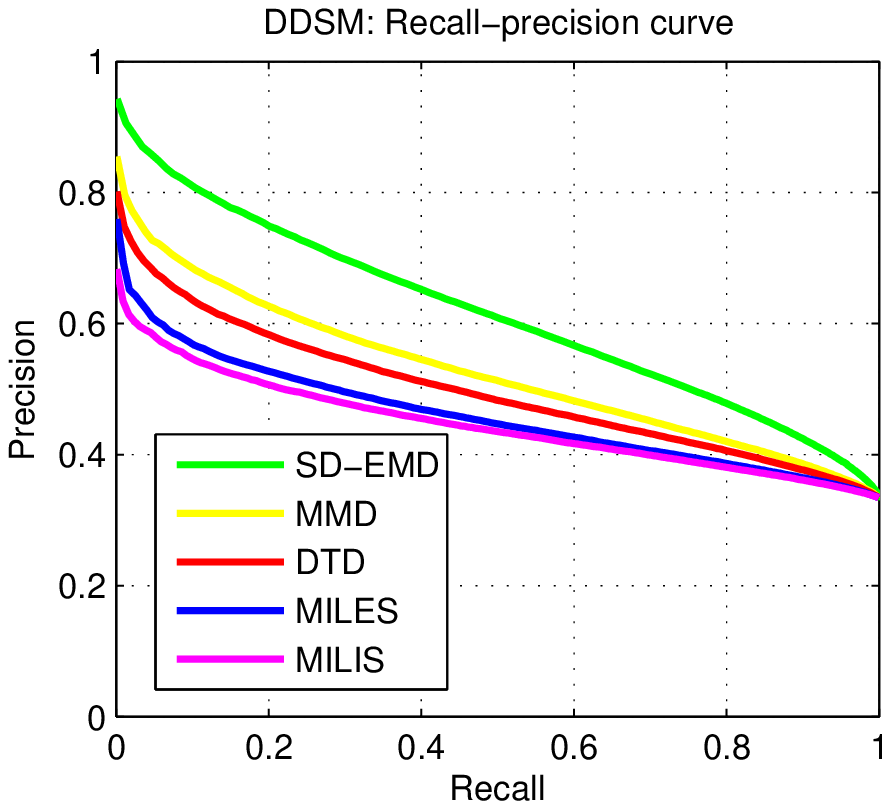}\\
\includegraphics[width=0.7\textwidth]{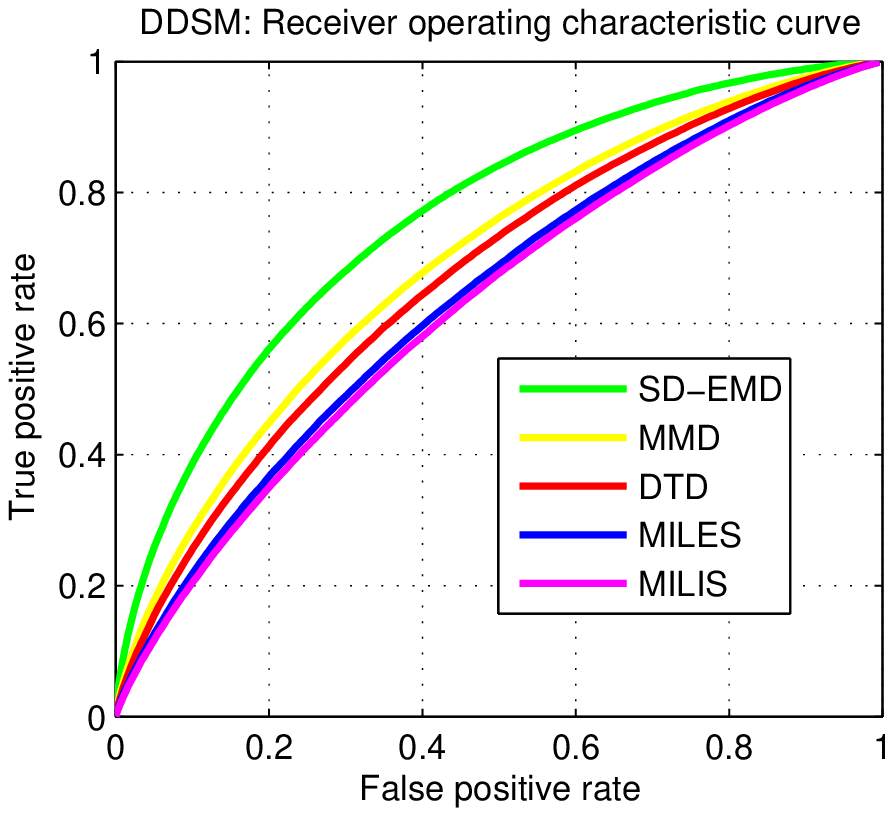}\\
\caption{Experimental results over the DDSM data set.}
\label{fig:DDSM}
\end{figure}

The experimental results of the compared methods over the  are given in the figure \ref{fig:DDSM}. We can clearly see that that the proposed method outperforms all the other methods significantly. As shown in the the recall-precision figure of the figure \ref{fig:DDSM}, the recall-precision curve of the proposed method, SD-EMD, is much more closer to the top-left corner than the other methods. This is not surprising because in the retrieval system, the EMD measure is used as distance measure, and only the SD-EMD method optimize the dictionary according to the EMD. The other methods ignores the EMD measure in their training process. From the receiver operating characteristic curve of the figure \ref{fig:DDSM}, it is also observed that the curve of SD-EMD is more closer to the top-right corner than all than other methods. This is strong evidence that the novel method works better than other methods.

\subsection{Results over the SemEval-2010 Task 8 data set}

\begin{figure}
\centering
\includegraphics[width=0.7\textwidth]{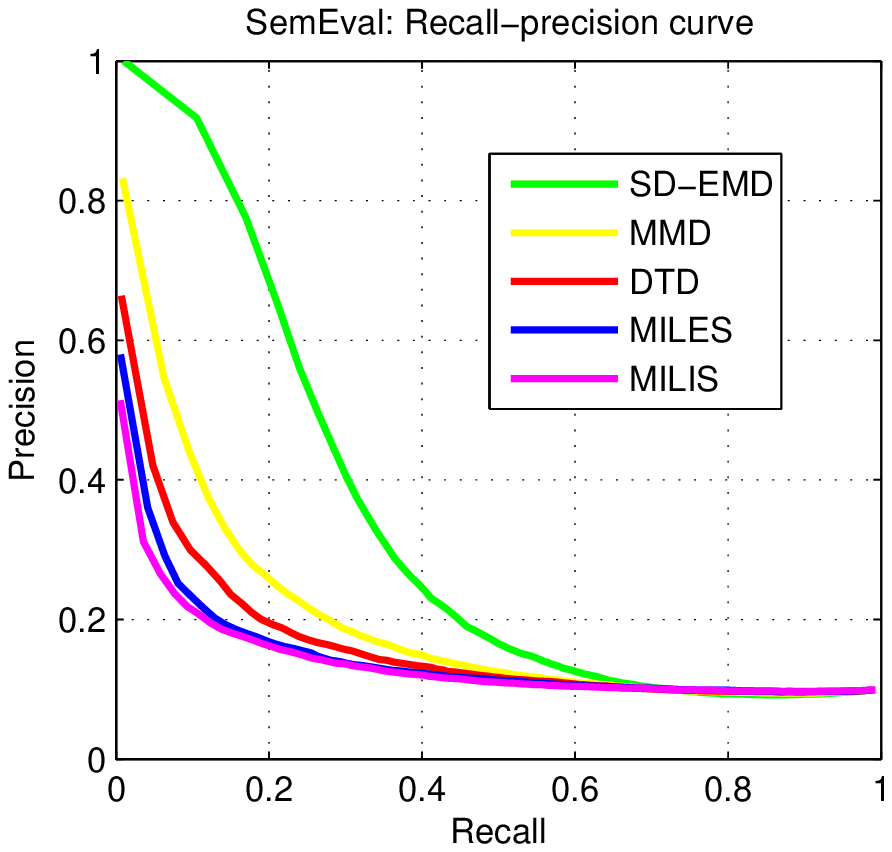}\\
\includegraphics[width=0.7\textwidth]{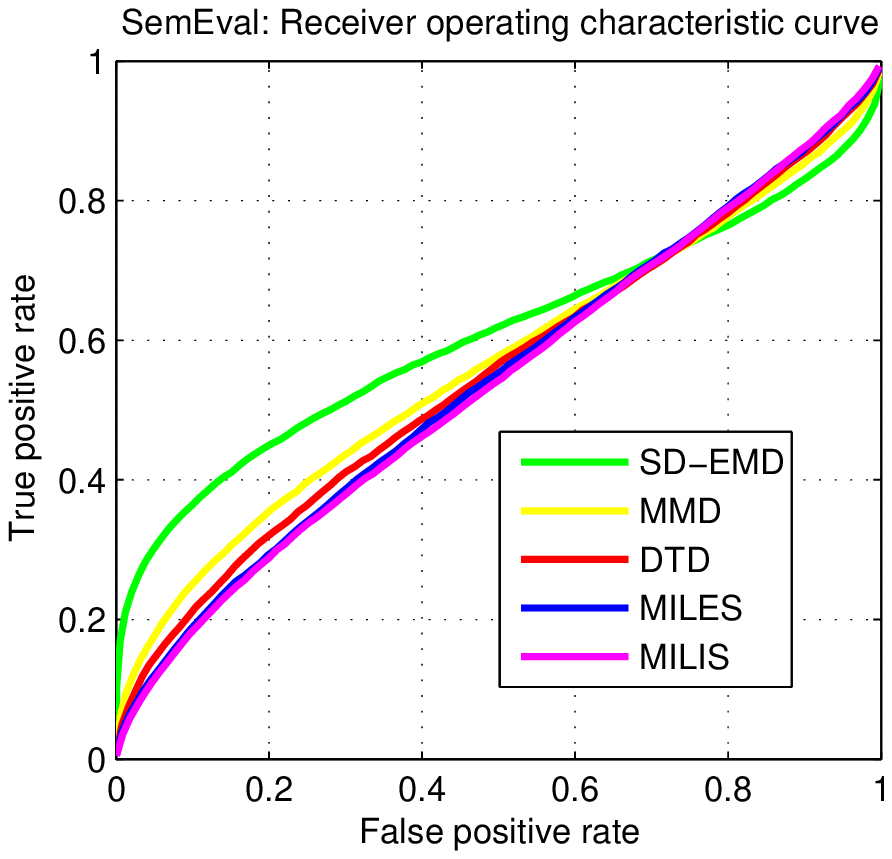}\\
\caption{Experimental results over the SemEval-2010 Task 8 data set.}
\label{fig:SemEval}
\end{figure}

The recall-precision curve and the receiver operating characteristic curve of the experiments over the SemEval-2010 Task 8 data set are shown in \ref{fig:SemEval}. Again, we observed that the proposed method, SD-EMD, outperforms all the compared methods over the SemEval-2010 Task 8 data set. This means that the proposed EMD based dictionary learning method not only works well over the medical image data, but also has good performances over the natural language processing problems.

\section{Conclusions}
\label{sec:conclu}

In this paper, we discuss the problem of representing multi-instance data sample by using the dictionary. The bags of instances are represented as histograms and the EMD is used to compare histograms. In this paper, for the first time, we proposed to learn EMD-optimal dictionary. We proposed a novel learning framework to train the dictionary by using the stochastic optimization method. The proposed method shows its advantage over the existing dictionary learning methods.

\section{Future works}

In this paper, we use the large margin as the loss function to model the problem. In the future, we will consider using more complex loss functions of multivariate performance measures \cite{liang2016optimizing,lin2016multi,wang2015multiple,fan2010enhanced}. Moreover, we will also consider using Bayesian network to represent the histograms and learn the parameters of Bayesian network for EMD based comparison \cite{fan2014tightening,fan2014finding,fan2015improved}. In the future, we will also applied the proposed method to various applications, including multimedia technology \cite{wang2014effective,wang2015supervised,liu2015supervised,liang2016novel}, information security \cite{xu2013cross,xu2014adaptive,xu2014evasion,xu2012push}, computer vision \cite{wang2015deeply,wang2013can,wang2013gender,wang2014leveraging}, medical imaging \cite{li2015outlier,li2015burn,mo2015importance,king2015surgical}, etc.

%\bibliographystyle{spmpsci}
%\bibliography{DictEMD}

\end{document}